\documentclass[nohyperref]{article}

\usepackage{microtype}

\usepackage{hyperref}

\usepackage[accepted]{arxiv}

\usepackage{amsmath}
\usepackage{amssymb}
\usepackage{mathtools}
\usepackage{amsthm}

\usepackage[capitalize,noabbrev]{cleveref}

\usepackage{graphicx}
\usepackage{amsmath}
\usepackage{amssymb}
\usepackage{booktabs}
 
\usepackage[utf8]{inputenc} 
\usepackage[T1]{fontenc}    
\usepackage{url}            
\usepackage{amsfonts}       
\usepackage{booktabs} 
\usepackage{nicefrac}       
\usepackage{microtype}      
\usepackage[export]{adjustbox}
\usepackage{bm}
\usepackage{xcolor}
\usepackage{amsmath}
\usepackage{common}
\usepackage{caption}
\usepackage{subcaption}
\usepackage{multirow}
\usepackage{nicefrac}
\usepackage{graphics}
\usepackage{xspace}
\usepackage{makecell}

\newcommand{\exalgo}{{\cal{A}}}
\newcommand{\lexp}{{\mathit{e}}}
\newcommand{\data}{{\cal{D}}}
\newcommand{\classexp}{{\cal{Y}}}
\newcommand{\Natural}{\mathbb{N}}

\newcommand{\pcl}{\theta^{CL}}

\newcommand{\hexp}{\bm{h}^{SC}}

\newcommand{\Dood}{\data_{ood}}

\newcommand{\fexp}{f^{SC}}

\newcommand{\kd}{{\cal{L}}^{KD}}

\newcommand{\loss}{{\cal{L}}}

\newcommand{\emptycell}{ \multicolumn{1}{c}{--} }

\theoremstyle{plain}

\theoremstyle{definition}

\theoremstyle{remark}

\icmltitlerunning{Projected Latent Distillation for Data-Agnostic Consolidation in Distributed Continual Learning}

\begin{document}

\twocolumn[
\icmltitle{Projected Latent Distillation for Data-Agnostic Consolidation in \\ Distributed Continual Learning}
%



\icmlsetsymbol{equal}{*}

\begin{icmlauthorlist}
\icmlauthor{Antonio Carta}{unipi}
\icmlauthor{Andrea Cossu}{sns}
\icmlauthor{Vincenzo Lomonaco}{unipi}
\icmlauthor{Davide Bacciu}{unipi}
\icmlauthor{Joost van de Weijer}{cvc}
\end{icmlauthorlist}

\icmlaffiliation{unipi}{Department of Computer Science, University of Pisa, Pisa, Italy}
\icmlaffiliation{sns}{Scuola Normale Superiore, Pisa, Italy}
\icmlaffiliation{cvc}{Computer Vision Center, Barcelona, Spain}

\icmlcorrespondingauthor{Antonio Carta}{antonio.carta@unipi.it}

\icmlkeywords{Continual Learning; Lifelong Learning; Communication-efficient learning}
\vskip 0.3in
]



\printAffiliationsAndNotice{}  

\begin{abstract}
    Distributed learning on the edge often comprises 
    \emph{self-centered devices} (SCD) which learn local tasks independently and are unwilling to contribute to the performance of other SDCs. \emph{How do we achieve forward transfer at zero cost for the single SCDs?} We formalize this problem as a \emph{Distributed Continual Learning} scenario, where SCD adapt to local tasks and a CL model consolidates the knowledge from the resulting stream of models without looking at the SCD's private data. 
    Unfortunately, current CL methods are not directly applicable to this scenario.
    We propose Data-Agnostic Consolidation (DAC), a novel double knowledge distillation method that consolidates the stream of SC models without using the original data. DAC performs distillation in the latent space via a novel Projected Latent Distillation loss. Experimental results show that DAC enables forward transfer between SCDs and reaches state-of-the-art accuracy on Split CIFAR100, CORe50 and Split TinyImageNet, both in reharsal-free and distributed CL scenarios. Somewhat surprisingly, even a single out-of-distribution image is sufficient as the only source of data during consolidation.
\end{abstract}

\section{Introduction}

In many machine learning applications, such as model personalization on edge devices, \emph{self-centered devices (SCDs)} train specialized models independently, each learning a subset of a larger problem. SCDs are "selfish actors" that do not want to share their private data or spend their limited computational and communication budget to support other SCDs. \emph{How can we improve the performance of SCDs at (almost) zero cost for them?}

To encourage positive transfer between the SCDs, we assume the existence of a consolidated model, which represents the knowledge learned by all the SCDs trained up to a certain moment in time. Each SCD will receive the current parameters of the consolidated model before training. In exchange, the SCD will send its trained model \emph{only once} after training. Training the consolidated model is a continual learning (CL) problem~\citep{lesortContinualLearningRobotics2020} because each model will arrive at different times and the consolidated model must be continuously updated. Unlike popular continual learning scenarios~\citep{vandevenThreeScenariosContinual2019}, the consolidated model does not have access to the original data but only to the SCD parameters.

In this paper, we formalize this problem as a \emph{Distributed Continual Learning} (DCL) scenario. Unfortunately, popular CL methods are unable to learn from stream of pretrained models without the original data. In fact, the use of pretrained models in most CL methods is limited to the parameters' initialization during the first step~\citep{hayesLifelongMachineLearning2020,maltoniContinuousLearningSingleincrementaltask2019}. This means that it is not possible to exploit any pretrained model that becomes available after training starts. 
Another challenge of our scenario is the strong asymmetry between SCDs and the consolidated model and the asynchronous nature of their communication. This is quite unlike federated learning~\citep{liFederatedLearningChallenges2020}, where a single model is trained in a centralized manner with multiple rounds of communication.

\begin{figure*}[t]
        \centering
        \includegraphics[width=0.7\textwidth]{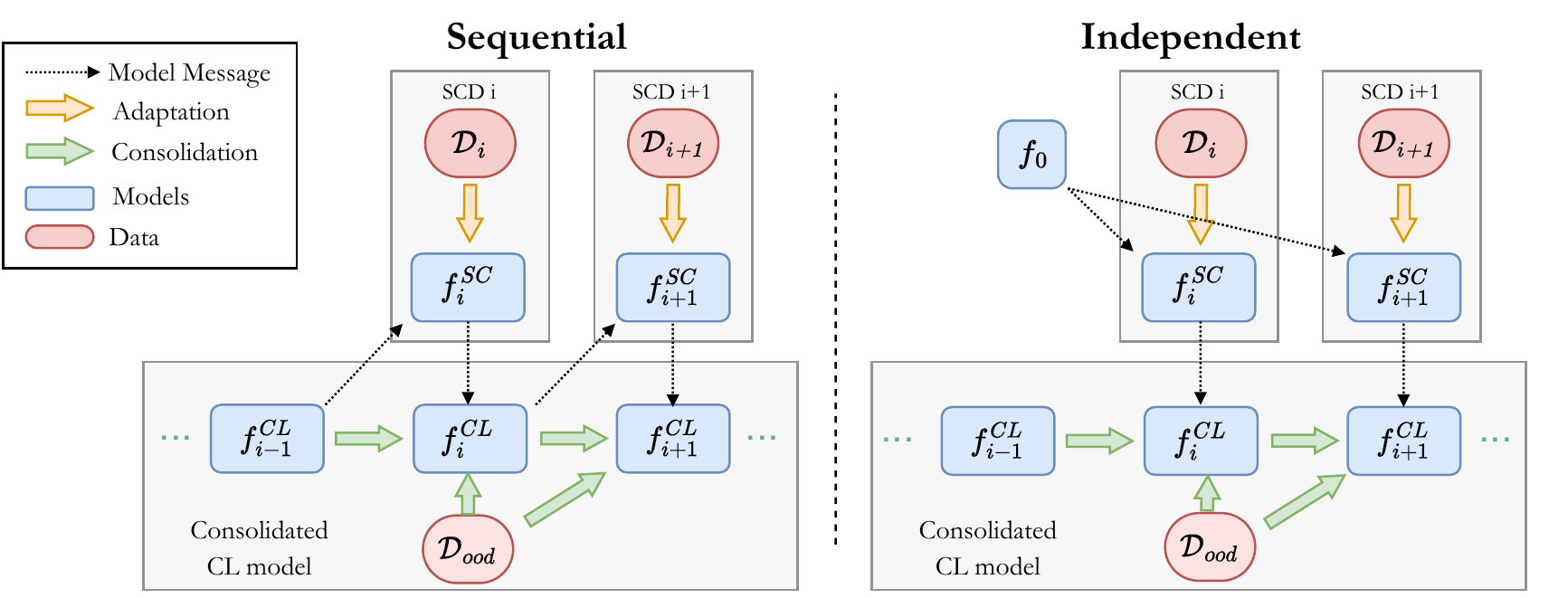}
        \caption{Sequential (left) and Independent (right) Distributed Continual Learning scenarios. SCDs (top row) learn single tasks and share the trained model with the consolidated CL model (bottom). Since the SCDs' data never leaves the device, the CL model uses only external data $\data_{ood}$ during training.}\label{fig:scenario}
\end{figure*}

We propose a novel method to train the consolidated CL model, called \emph{Data-Agnostic Consolidation (DAC)}, that performs double knowledge distillation from the previous consolidated and current self-centered model without using the original data. DAC can consolidate the two models using a single out-of-distribution image and heavy augmentations, in line with recent advances for knowledge distillation~\citep{beyerKnowledgeDistillationGood2022,asanoExtrapolatingSingleImage2022}. As a consequence, DAC can iteratively update the consolidated model without ever looking at the original data. 
A limitation of current double distillation is that it is not possible to perform feature distillation because the two teachers compute two different latent representations. DAC solves this problem via our novel \emph{Projected Latent Distillation} loss. Since DAC uses the same loss for the old and new model, it controls the stability-plasticity tradeoff~\citep{frenchCatastrophicForgettingConnectionist1999}, resulting in good performance even for old tasks, something that most methods struggle with. The experimental results show that DAC allows to consolidate the knowledge from independent agents even when they are trained on different tasks and only a single out-of-distribution image is available.

The main contributions of the paper can be summarized as follows:
\begin{itemize}
    \item a formal characterization of distributed continual learning. To the best of our knowledge, this is the first work to formalize continual learning in a distributed setting and the consolidation problem (Section \ref{sec:scenario});
    \item \emph{Data-Agnostic Consolidation}, a novel strategy which performs a data-agnostic double knowledge distillation in the output and latent space via Projected Latent Distillation (Section \ref{sec:method});
    \item state-of-the-art results in the classic rehearsal-free and the novel DCL scenario for task-aware SplitCIFAR100 (Table \ref{tbl:cifar100}, +3.9\% on 10 Tasks), task-aware Split TinyImageNet, and task-agnostic CORe50-NI\footnote{\url{https://github.com/AntonioCarta/data-agnostic-consolidation}};
    \item we show that DAC allows positive forward transfer between the SCDs and that simple sources of data are sufficient for consolidation when coupled with heavy augmentations (Section \ref{sec:experiments}).
\end{itemize}

\section{Distributed Continual Learning}\label{sec:scenario}
In Distributed Continual Learning (DCL) we have two types of models interacting with each other: \emph{self-centered models}, denoted as $f^{SC}$, and the \emph{consolidated} model, denoted as $f^{CL}$. SC models perform a local knowledge adaptation step by learning a single task, while the CL model consolidates the knowledge of a stream of SC models. 
SCDs have four characteristics: (i) they want to train independently, (ii) they never share the data, (iii) they want minimal communication, and (iv) they do not care about forgetting other tasks.
Given these constraints, we define knowledge adaptation as the local problem solved by each SCD, and knowledge consolidation as the continual learning problem where a model must iteratively consolidate SC models without access to the original data. More formally:
\paragraph{Knowledge Adaptation} is an algorithm $\exalgo^{ADA}:\ \langle f^{0}_{i-1}, \data^i_{train}\rangle\ \rightarrow\ \langle \fexp_i\rangle$ that takes an initialization $f^{0}_{i-1}$ and a batch of data $\data^i_{train}$ and returns a model $\fexp_i$. This is a simple finetuning procedure where only the local loss $\loss(\fexp_i, \data_i)$ is minimized. We call the resulting model $\fexp_i$ the self-centered model for task $i$.
\paragraph{Knowledge Consolidation} is an algorithm $\exalgo^{KC}:\ \langle f^{CL}_{i-1}, \fexp_i, \data^{ood}\rangle\ \rightarrow\ \langle f^{CL}_{i} \rangle$ that takes as input two models, the previous CL model $f^{CL}_{i-1}$ and the current SC model $\fexp_{i}$, and a small dataset $\data^{ood}$ and combines them into a single model $f^{CL}_{i}$. The resulting model is optimized to solve the task encountered by both input models. The dataset $\data^{ood}$ used during the consolidation is different from the original data used to train the SC models.

To ensure a fixed memory occupation we assume that a knowledge consolidation algorithm can store only two models at the same time, the current SC model and the previous consolidation model.

\subsubsection*{Two Minimal Communication Schemes: The Sequential and Independent Settings}
The communication scheme between the consolidated and self-centered models, shown in Figure \ref{fig:scenario} consists two messages: the \emph{initialization message}, sent from the consolidated model to the SCD before training, and the \emph{SC message}, sent from the SCD to the consolidated model after training. The communication happens asynchronously only at the beginning and at the end of training for each SCD, allowing complete freedom during training. The initialization message allows positive transfer between the SCDs, while the SC message allows to update the CL model with the new task learned by the SC.

The interaction between the SC and CL model happens in this order: (1) Before training, the SCD receives its initialization from the CL model. (2) The SCD performs a \emph{knowledge adaptation} step to learn its task. (3) After training, the SCD sends its trained model to the CL model. (4) The CL model performs a \emph{knowledge consolidation} step to integrate the local model into its global model. 

Under this communication scheme, the CL model receives a sequence of expert models $\fexp_0, \hdots, \fexp_i$ and aggregates them via a knowledge consolidation algorithm. Notice that the global model never receives the original data.
During this interaction, the ordering between the messages arriving the SCDs is critical since it determines how much transfer there can be between the SC models. We focus on two extremes of the range of possibilities: the \emph{sequential} and \emph{independent} scenario, which correspond to the consolidation of sequential or completely independent SC models, respectively.

In the \emph{sequential scenario} (Fig \ref{fig:scenario}, left), SC models are trained sequentially. At time $i$, the SCD $i$ receives $f^{CL}_{i-1}$ from the CL model. The next SCD ($i+1$) will start training only after the consolidation of the SC model $\fexp_i$. This is the scenario where there is the opportunity for transfer.

In the \emph{independent scenario} (Fig \ref{fig:scenario}, right), all the SCDs start training in parallel at time $t=0$. Therefore, they receive the same random initialization $f_0$. In this setting, there is no transfer between the SC models.

We use the two scenarios to evaluate the impact of knowledge transfer between between SCDs (Section \ref{sec:cka_forward}).

\subsubsection*{Relationship with Classic Continual Learning}
Continual learning is the problem of learning from a non-stationary stream of experiences $S = \lexp_1, \hdots, \lexp_n$. In classification problems, each experience $\lexp_i$ provides a batch of samples $\data_i = \{ \langle x_m, y_m, t_m \rangle \}$, where $x_m \in \Real^X$ is the input, $y_m \in \classexp_i$ the target label, and $t_m \in \Natural$ an optional task label. A continual learning model $f^{CL}$ learns on each experience sequentially. In the rehearsal-free setting at time $i$ data $\data_j, j<i$ from previous experiences is not available~\citep{smithAlwaysBeDreaming2021,masanaClassIncrementalLearningSurvey2022}. A rehearsal-free continual learning algorithm $\exalgo^{CL}:\ \langle f^{CL}_{i-1}, \data^i_{train}\rangle\ \rightarrow\ \langle f^{CL}_i\rangle$ takes as input the previous model $f^{CL}_{i-1}$ and the new data $\data^i_{train}$ and returns the updated model $f^{CL}_i$. Each experience $k$ has a separate loss $\loss^{i,k} = \loss(f^{CL}_i, \data_k)$ and the objective of the algorithm is to minimize the loss over the entire stream $\loss^S = \sum_{k=1}^n \loss^{n,k}$.

In general, DCL is more constrained than other CL scenario, even rehearsal-free scenarios since the consolidated model will never see any real data. As a result, most CL strategies cannot be applied to DCL. Conversely, DCL methods can be applied to rehearsal-free CL by performing the adaptation followed by the consolidation. In fact, in our experiments we show that our proposed method is competitive even in a classic rehearsal-free scenario (Section \ref{sec:experiments}).

\section{Data-Agnostic Consolidation}\label{sec:method}

\begin{figure*}
    \centering
    \includegraphics[width=0.8\textwidth]{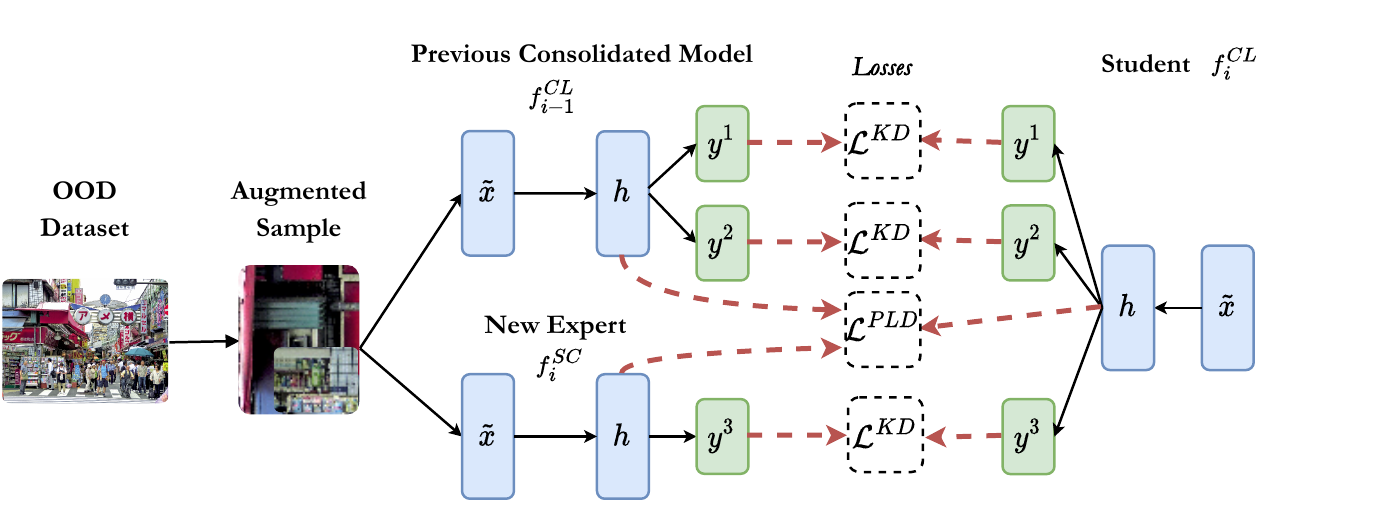}
    \caption{Data-Agnostic Consolidation. The method uses double distillation in the output and latent space, using heavily augmented samples as input.}\label{fig:pld}
\end{figure*}

As discussed above, DCL requires two algorithms: knowledge adaptation and knowledge consolidation. Since knowledge adaptation aims for optimal plasticity in the SCD task, we solve it by finetuning the initialization received by the CL model to minimize the loss on the local data $\data_i$ via stochastic gradient descent.
To solve knowledge consolidation we propose \emph{Data-Agnostic Consolidation (DAC)}, which we describe below. DAC is a method for knowledge consolidation that matches the output and feature space of the two models via a double knowledge distillation (Section \ref{sec:fun_matching}). At each iteration, DAC samples from a source of data and uses heavy augmentations to increase the data diversity (Section \ref{sec:dac_data}). Finally, DAC leverages a novel approach called \emph{Projected Latent Distillation (PLD)} to distill the latent spaces of the two models (Section \ref{sec:dac_pld}). While our main objective with DAC is to solve DCL, the method can be also be used in rehearsal-free CL (shown in Section \ref{sec:experiments}).

\subsection{Knowledge Consolidation Objective}\label{sec:fun_matching}

In knowledge consolidation, we combine the previous CL model $f^{CL}_{i-1}$ and the current SC mdoel $\fexp_i$. The resulting model $f^{CL}_i$ consolidates the knowledge of both models. As discussed before, we do not have access to $\data_i$ during the consolidation. Therefore, we use an out-of-distribution dataset $\data_{ood}$.

We assume that $f^{CL}_i$, the current CL model, is a multi-head model, with a separate linear head for each task $k$. We denote $f^{CL}_i(\bm{x}, k)$ as the output for the head $k$ corresponding to input $\bm{x}$. Since DAC uses multi-head models during consolidation, during inference we select the correct head if task labels are available. In task-agnostic scenarios, we average (or concatenate, if they predict different classes) the outputs of all the heads.

At step $i$, we want to consolidate a multi-head model $f^{CL}_{i-1}$ with $i-1$ heads and a single-head model $\fexp_i$. The desired result is a multi-head model $f^{CL}_{i}$ with $i$ heads such that
\begin{align}
    f^{CL}_i(\bm{x}, k) &= f^{CL}_{i-1}(\bm{x}, k), 
        \quad \forall \bm{x} \in \mathbb{R}^N, k \neq i\\
    f^{CL}_{i}(\bm{x}, i) &= \fexp_{i}(\bm{x}), 
        \quad \forall \bm{x} \in \mathbb{R}^N \label{eq:funmatching}.
\end{align}
Notice that, since we want to match the two teachers exactly on the entire input space, this is an exact optimization objective. In particular, its definition does not depend on the original data used to train the teachers. We can optimize $\pcl_i$ by stochastic gradient descent minimizing a double knowledge distillation loss
\begin{align}
    E_{\bm{x} \sim \data^{ood}}[\loss^{DKD}(\bm{x})] =& \loss^{KD}(f^{CL}_i(\bm{x}, i), \fexp_i(\bm{x})) \\
        + \sum_{k=1}^{i-1} \loss^{KD}(& f^{CL}_i(\bm{x}, k), f^{CL}_{i-1}(\bm{x}, k)) \label{eq:lossddoublekd}
\end{align}
where $\kd(\bm{y}_S, \bm{y}_T)$ is the KL-divergence between the student's output $\bm{y}_S$ and the teacher's output $\bm{y}_T$. $\loss^{DKD}$ balances each task contribution by summing the head's errors. Since we don't have access to the original data and the objective in Eq. \ref{eq:funmatching} does not depend on the original data, we need to provide an alternative source of data $\data^{ood}$.

\subsection{Sampling Data for Consolidation}\label{sec:dac_data}
DAC is agnostic to the data used for consolidation. If available, it is possible to use the original data. However, in DCL we assume to only have access to a small set of samples $\Dood$. In our experiments, we have found a single image to be sufficient in many experimental settings. Following \citet{asanoExtrapolatingSingleImage2022} and \citet{beyerKnowledgeDistillationGood2022}, we use a large set of stochastic augmentations to create a dataset of highly diverse samples from a small number of images. At each timestep, we apply a large number of transformations such as jittering, rotation, crop, resize, flip, CutMix~\citep{yunCutMixRegularizationStrategy2019a}. The resulting images will be heavily distorted but we can still use them to match the output targets of the two teachers in Eq. \ref{eq:funmatching}. Since we don't have the original images, augmentations ensure that we have enough diversity in our data, which as we will show in Section \ref{sec:experiments} is critical for success. While the preprocessing pipeline can become the bottleneck of the training process with such a large number of augmentations, it is always possible to trade-off memory to save computation by precomputing a large number of pre-processed images.


\begin{table*}
    \scalebox{0.85}{
        \begin{subtable}[h]{.38\linewidth}
            \centering
            \captionof{table}{Task-incremental SplitCIFAR100 after task 5 and 10. Baselines denoted by $\dagger$ are taken from \cite{masanaClassIncrementalLearningSurvey2022}}\label{tbl:cifar100}
            \begin{tabular}{lcrr}
            \toprule
                    & \multicolumn{3}{c}{SplitCIFAR100} \\
                    & DCL & 5 Tasks & 10 Tasks \\ \midrule
            Naive$^\dagger$   & RF & 49.8              & 38.3               \\
            EWC$^\dagger$     & RF & 60.2              & 56.7               \\
            PathInt$^\dagger$ & RF & 57.3              & 53.1               \\
            MAS$^\dagger$     & RF & 61.8              & 58.6               \\
            RWalk$^\dagger$   & RF & 56.3              & 49.3               \\
            LwM$^\dagger$     & RF & 76.2              & 70.4               \\
            LwF$^\dagger$     & RF & 76.7              & 76.6               \\ 
            DMC$^\dagger$     & \checkmark & 72.3              & 66.7               \\
            \midrule
            DAC($\lambda=0$)  & \checkmark & \mustd{77.6}{1.7}          & \mustd{77.5}{0.6} \\
            DAC               & \checkmark & \textbf{\mustd{81.4}{1.6}} & \textbf{\mustd{80.5}{0.8}}  \\ \bottomrule 
            \end{tabular}
        \end{subtable}%
    }
    \hspace{1em}%
    \scalebox{0.85}{
        \begin{subtable}[h]{.3\linewidth}
            \centering
            \captionof{table}{Task-incremental Split Tiny ImageNet. Baselines denoted by $\dagger$ are taken from \cite{delangeContinualLearningSurvey2022}}\label{tbl:simnet}
            \begin{tabular}{lcr}
            \toprule
                    \multicolumn{3}{c}{Split Tiny ImageNet} \\
                    & DCL & ACC \\
            \midrule
            Joint$^\dagger$      & B & 57.29  \\
            Naive$^\dagger$      & RF & 25.28 \\
            PackNet$^\dagger$    & RF & 47.64 \\
            HAT$^\dagger$        & RF & 43.78 \\
            SI$^\dagger$         & RF & 33.86 \\
            EWC$^\dagger$        & RF & 31.10 \\
            MAS$^\dagger$        & RF & 45.08 \\
            LwF$^\dagger$        & RF & 46.79 \\
            EBLL$^\dagger$       & RF & 46.25 \\
            mode-IMM$^\dagger$   & RF & 36.42 \\ 
            \midrule
            DAC($\lambda = 0$)   & \checkmark & \mustd{40.46}{9.2}  \\ 
            DAC                  & \checkmark & \textbf{\mustd{48.3}{0.5}}  \\ 
            \bottomrule
            \end{tabular}
        \end{subtable}%
    }
    \hspace{1em}%
    \scalebox{0.85}{
        \begin{subtable}[h]{.32\linewidth}
            \caption{CORe50 task-agnostic independent DCL scenario. Baselines denoted by $\dagger$ are taken from \cite{cartaExModelContinualLearning2022}.}
            \label{tab:res_core50}
            \begin{tabular}{lcrrrrr}
                \toprule
                                    & & \multicolumn{2}{c}{CORe50} \\
                                    & DCL & Joint & NI \\ \midrule
                Oracle$^\dagger$              & B & \mustd{85.7}{0.2} & \emptycell \\ 
                Min. Entropy$^\dagger$        & B & \emptycell          & \mustd{61.3}{1.8} \\
                Output Avg.$^\dagger$         & B & \emptycell          & \mustd{69.9}{0.7} \\
                Replay ED$^\dagger$           & B & \mustd{87.4}{0.2} & \mustd{83.7}{0.5} \\ \midrule
                Parameter Avg.$^\dagger$      & \checkmark & \emptycell          & \mustd{2.0}{0.0} \\ \midrule
                MI-ED$^\dagger$  & \checkmark & \mustd{50.0}{2.7} & \mustd{44.3}{4.9} \\
                DI-ED$^\dagger$  & \checkmark & \mustd{52.9}{2.0} & \mustd{43.2}{2.3} \\
                Aux. Data ED$^\dagger$        & \checkmark & \mustd{81.8}{0.2} & \mustd{44.5}{2.9} \\ \midrule
                DAC($\lambda=0$)     & \checkmark & \mustd{82.2}{0.2} &  \mustd{41.5}{9.07} \\
                DAC                  & \checkmark & \mustd{84.9}{0.2} & \mustd{48.9}{1.9} \\
                \bottomrule
            \end{tabular}
        \end{subtable}
    }
    \caption{Average test set accuracy $A_{t}$. The column DCL shows whether a strategy satisfies the DCL constraints (\checkmark), it is a rehearsal-free strategy (RF), or it is a baselines which uses replay data or keeps multiple models in memory (B).}
\end{table*}


\subsection{Projected Latent Distillation}\label{sec:dac_pld}
The knowledge distillation loss in Eq. \ref{eq:lossddoublekd} matches the outputs for each teacher's head. However, we would like to also match the latent space of the two teachers. This is more problematic because given an input $\bm{x}$, while the outputs $\bm{y}_k$ are computed by separate heads, and therefore have no interference, the hidden activations share the same units in the consolidated model (as shown in Figure \ref{fig:pld}).
As a result, for a specific hidden layer we have two different targets $\bm{h}^{CL}_{{i-1}}$ and $\hexp_{i}$ and a single activation vector $\bm{h}^{CL}_{i}$ for the student. Exact matching of the hidden state is not possible. To solve this issue, we propose \emph{Projected Latent Distillation (PLD)}. 
The underlying intuition is that while we cannot enforce an exact match, we can match the two hidden states up to a linear transformation. During the consolidation phase, we optimize two linear transformations $\bm{W}^{SC}$ and $\bm{W}^{CL}$ that map the teachers' hidden states to the student's hidden states. The loss is then defined as
\begin{align}
    \loss^{PLD}(\bm{h}^{CL}_i, \bm{h}^{CL}_{i-1}, \hexp_{i}) = \lambda ( 
        &|| \bm{W}^{SC} \bm{h}^{CL}_i - \hexp_{i} ||^2_2 \\ 
        + (i-1) &|| \bm{W}^{CL} \bm{h}^{CL}_i - \bm{h}^{CL}_{i-1} ||^2_2 ), \label{eq:lossdpl}
\end{align}
where $\bm{h}^{CL}_i$ is the student, $\bm{h}^{CL}_{i-1}$ the previous CL model (already trained on $i-1$ tasks) and $\hexp_{i}$ the SC model's hidden states. The linear transformations are initialized to the identity matrix and optimized during the consolidation phase. The loss encourages the student model to also match the teachers' hidden states. Notice that the loss for $f^{CL}_{i-1}$ is multiplied by $i-1$ to give the same weight to each task. The same effect is present in the distillation in the output space defined by Eq. \ref{eq:lossddoublekd}, since we sum all the heads (one for each task). The total loss of DAC is $\loss(\bm{x}, \bm{h}^{CL}_i, \bm{h}^{CL}_{i-1}, \hexp_{i}) =  \loss^{DKD}(\bm{x}) + \lambda \loss^{PLD}(\bm{h}^{CL}_i, \bm{h}^{CL}_{i-1}, \hexp_{i})$, where $\lambda$ controls the ratio between the ouput and latent distillation losses.

A schematic view of DAC is shown in Figure \ref{fig:pld}.


\section{Experiments}\label{sec:experiments}

We show experimental results in the sequential and independent setting introduced in Section \ref{sec:scenario}. We use CIFAR100~\citep{krizhevskyLearningMultipleLayers2009}, Tiny ImageNet~\citep{chrabaszczDownsampledVariantImageNet2017} and CORe50 ~\citep{pmlr-v78-lomonaco17a} to create our benchmarks. The source code is implemented in Avalanche~\citep{lomonacoAvalancheEndtoEndLibrary2021} and publicly available\footnote{\url{https://github.com/AntonioCarta/data-agnostic-consolidation}}. More details about the experiments can be found in the appendix.

We use SplitCIFAR100 in the task-incremental setting using 10 and 20 tasks, where the stream is divided into experiences of 10 and 5 classes, respectively. We use a slimmed ResNet18 as defined in ~\citep{lopez-pazGradientEpisodicMemory2017}.
We also use Split Tiny ImageNet (10 Task) in the task-incremental setting with a WideVGG architecture following~\citet{delangeContinualLearningSurvey2022}.
Finally, we use CORe50~\citep{pmlr-v78-lomonaco17a}, a task agnostic-benchmark, in the joint and domain-incremental (NI) settings. CORe50 images are scaled to a $128 \times 128$ resolution. We use a MobileNetv2~\citep{howardMobileNetsEfficientConvolutional2017} pretrained on ImageNet, as it is popular in the literature~\citep{maltoniContinuousLearningSingleincrementaltask2019}.
We use these streams in a DCL scenario by giving each experience to a different SCD. SplitCIFAR100 and Split Tiny ImageNet are sequential DCL scenario, while CoRE50 is an independent DCL scenario.
The default data source for DAC is a single image ("animals", shown in the Appendix). 

\subsubsection*{Distributed Continual Learning Evaluation}
We point out that many strategies that we compare against do not satisfy the DCL constraints. In the following results, we highlight whether a strategy satisfies DCL constraints, rehearsal-free constraints, or it is simply a baselines that uses rehearsal data or keeps all the teachers in memory. The comparison against rehearsal-free baselines is used to show that DAC can also be applied in a rehearsal-free setting.
All the results are the average of 5 runs on different seeds.

The consolidated CL model is evaluated with the average accuracy over the entire test stream. Given $A_{t,i}$ as the task $i$ accuracy for task $i$ after training on task $t$, the average accuracy $A_t = \sum_{i=1}^t A_{t,i}$. This choice also allows us to compare DAC in a more classic rehearsal-free scenario.
We also evaluate the SC models by measuring their local task accuracy (Section \ref{sec:cka_forward}) and linear probing of the last layer to evaluate the finetuning performance on all the tasks.

\subsection{Rehearsal-Free and Distributed Continual Learning}

In this section we show how DAC performs in DCL scenarios. We also show the performance of state-of-the-art rehearsal-free baselines. Although these are not applicable to DCL, the results suggest that DAC is a good choice even in the more popular rehersal-free setting.
Results for SplitCIFAR100 in the task-aware sequential setting are shown in table \ref{tbl:cifar100}. We use the results in \citep{masanaClassIncrementalLearningSurvey2022} as baselines. DAC shows state of the art performance on the 10 task setting. Despite the limited data source, a single out-of-domain image ("animals"), DAC outperforms LwF, which uses the real data $\data_i$ for distillation. We argue that there are two properties of DAC which justify this improvement. First, the use of heavy augmentations improves distillation even with limited data, as already shown in \citep{beyerKnowledgeDistillationGood2022} and \citep{asanoExtrapolatingSingleImage2022} for offline training. Furthermore, during the consolidation DAC weighs the current task and all the previous ones in the same way by using the same loss for all tasks and summing them (Eq. \ref{eq:lossddoublekd}, \ref{eq:lossdpl}). Instead, LwF uses the cross-entropy for the current task and the KL divergence for the previous ones, which makes it more difficult to balance stability and plasticity (the well known stability-plasticity dilemma, ~\citep{frenchCatastrophicForgettingConnectionist1999}). We evaluated DAC on the 20 task setting (5 classes per task), which results in the average test accuracy of \mustd{86.2}{1.0}, even higher than the 10 task setting, suggesting that DAC can scale well to longer streams. Results on Tiny ImageNet in the sequential setting (Table \ref{tbl:simnet}) are consistent with Split CIFAR100, with DAC as the best performing method with a lower performance when PLD is not used ($\lambda=0$).

Experiments on CORe50 in the independent setting are shown in Table \ref{tab:res_core50}. Notice that while our method uses a multi-head, CORe50 is a task-agnostic benchmark. Therefore, we convert the final multi-head model into a task-agnostic model by averaging the output of all the heads. CORe50 is also a more challenging benchmark due to the higher image resolution ($128 \times 128$). We compare against the methods in \citet{cartaExModelContinualLearning2022}, which perform knowledge distillation using synthetic data or the entire ImageNet dataset. Overall, DAC obtains the best performance, even outperforming Aux. Data ED, which uses ImageNet (with more than 1 milion images) for distillation.

\begin{figure}
    \begin{subfigure}{0.45\textwidth}
        \centering
        \includegraphics[width=0.95\textwidth]{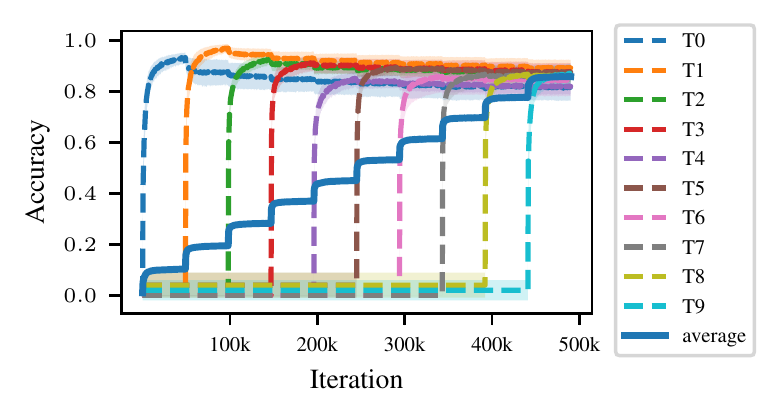}
        \caption{SplitCIFAR100.}
    \end{subfigure}
    \begin{subfigure}{0.45\textwidth}
        \centering
        \includegraphics[width=0.95\textwidth]{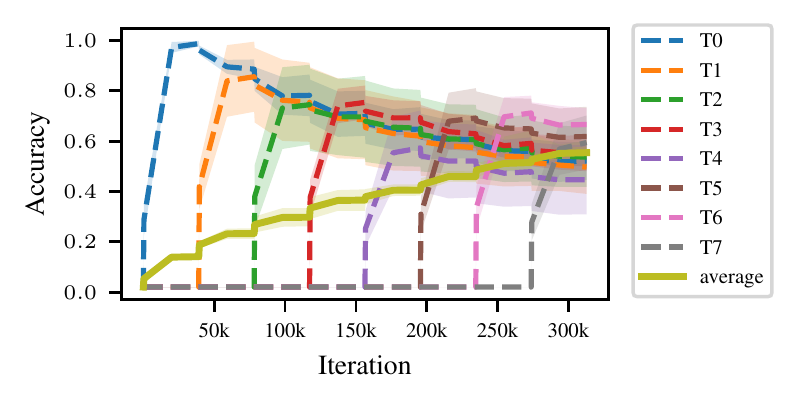}
        \caption{CORe50-NI.}
    \end{subfigure}
    \caption{Task-specific and stream average accuracy on the training set during training.}\label{fig:curves}
\end{figure}

Figure \ref{fig:curves} shows the learning curves for SplitCIFAR100 (10 Tasks) and CORe50-NI. We notice that on Split CIFAR100 the forgetting, i.e. the difference between the accuracy after training on task $i$ and the task at the end of training on the entire stream, is very low. This means that the consolidation process works with minimal forgetting. On CORe50, we see a bigger drop over time, but the old and new tasks have a similar performance. This suggests that the drop may be the result of the problem becoming more difficult with the introduction of new instances while keeping a fixed capacity.

\subsection{PLD Improves Forward Transfer}\label{sec:cka_forward}
\newcommand{\nopld}{\texttt{DAC($\lambda=0$)}}
\newcommand{\expdac}{\texttt{DAC}}

\begin{figure*}
    \centering
    \begin{subfigure}{.33\textwidth}
        \centering
        \includegraphics*[width=0.95\textwidth]{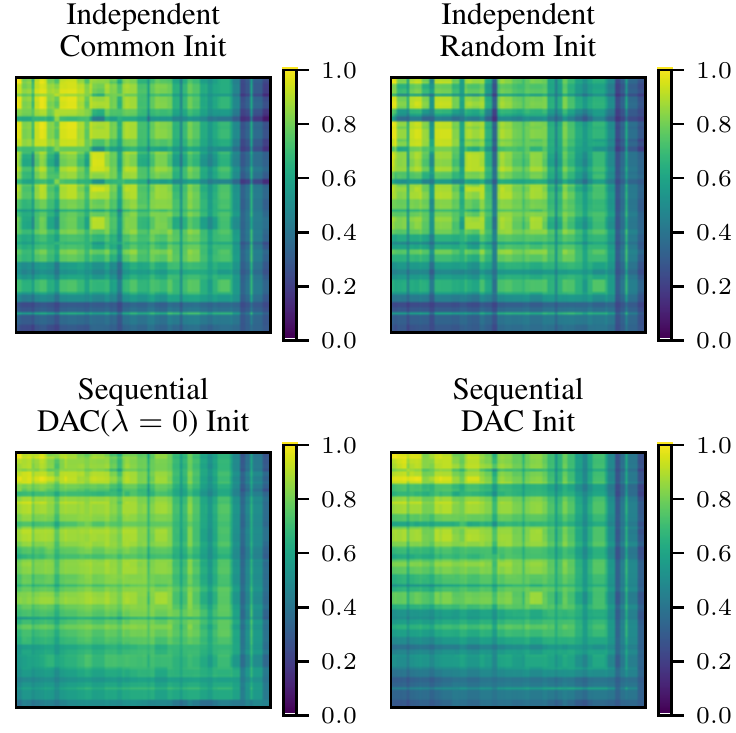}
        \caption{CKA between the first and last SC model of SplitCIFAR100 (10 Tasks) computed using the last task.}\label{fig:cka}
    \end{subfigure}
    \begin{subfigure}{.33\textwidth}
        \centering
        \includegraphics[width=\linewidth]{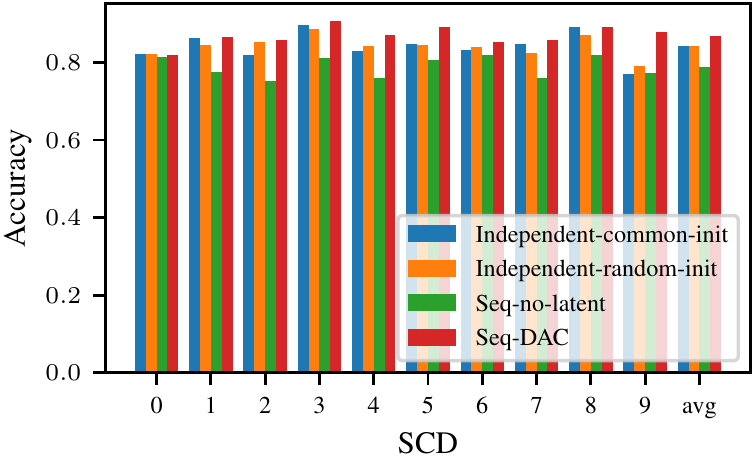}
        \caption{Accuracy on the SCD's task.}\label{fig:forward_curr}
    \end{subfigure}%
    \begin{subfigure}{.33\textwidth}
        \centering
        \includegraphics[width=\linewidth]{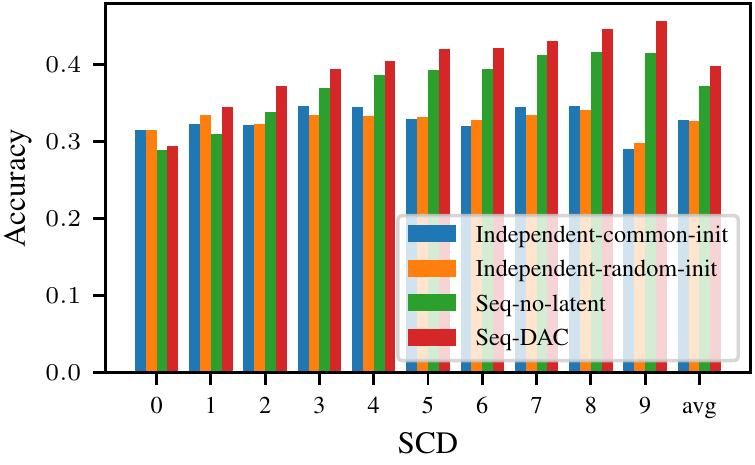}
        \caption{Task-agnostic linear probing on all tasks.}\label{fig:forward_all}
    \end{subfigure}
    \caption{Accuracy of the SC models on their own task and linear probing of the model's representation finetuned on the entire CIFAR100 dataset. Notice that PLD is not applied if $\lambda=0$.}
    \label{fig:forward_transfer}
\end{figure*}

The sequential and independent scenario differ only in the SC model's initialization. In the sequential setting, each agent is initialized with the weights learned at time $t-1$, while in the independent setting all the agents start from a common initialization $\theta^0$. The model's initialization affects the similarity between the SC models and therefore the difficulty of the consolidation problem. Sequential training also allows forward transfer between the SC models, which is not possible in the independent scenario.

In this section, we study how the two settings affect the SC model's accuracy with respect to 3 dimensions: forward transfer, the generalization of the hidden representations to other tasks, and the representation similarity between different SC models. We also ablate the use of Projected Latent Distillation for DAC to investigate its effect. We use SC models trained on SplitCIFAR100 (10 Tasks). We use a common or different random initialization for the SC models in the independent setting. In the sequential setting we compare DAC without latent distillation (\nopld), and the full DAC.

The sequential initialization should encourage forward transfer between the CL model and the SC model, i.e. it should improve the SC model's performance compared to the independent setting. Figure \ref{fig:forward_transfer} shows the accuracy of the 10 SC models on the task they have seen during training (left) and the average accuracy over all tasks for a task-agnostic linear probe which uses the final layer's representation, finetuned on all the tasks (right). Surprisingly, while \nopld\ SC models obtain a higher accuracy than independent SC models on the linear probing, the average SC model's accuracy on the task seen during training is actually lower. Therefore, it seems that without latent distillation there is negative forward transfer. Instead, \expdac\ SC models are better than the independent SC models, which means that the initialization favors a positive forward transfer. The linear probe accuracy (Fig. \ref{fig:forward_all}) on all tasks shows positive forward transfer for both \nopld\ and \expdac, albeit \expdac\ has a better performance.

In Figure \ref{fig:cka}, we measure representation similarity with the CKA \citep{nguyenWideDeepNetworks2022}. The figure shows the CKA between the first (Task 0) and last (Task 9) SC models\footnote{The CKA for the entire stream is shown in the appendix.}. In general, the CKA similarity is relatively high for the different configurations since they all share the same architecture and are trained on similar data. Somewhat surprisingly, the similarity between \expdac\ SC models is lower than the similarity of \nopld\ SC models. In principle, we expected to find a positive correlation between the representation similarity and the forward transfer. This appears not to be the case since \expdac\ has better forward transfer but \nopld\ shows closer similarity. We hypothesize that the PLD loss encourages the consolidated model to learn more diverse representations, decreasing the representation similarity over time but increasing the forward transfer thanks to the richer representation.

\subsection{Comparison Between Different Data Sources}\label{sec:data_ablation}
We have already shown that a single image is already competitive with state of the art methods. In this section, we study how different data source properties help the consolidation process (Table \ref{tbl:cifar100Abl}). We use data sources taken from the real stream, single images vs full dataset, and natural images against other domains, synthetic images and static noise:
\begin{itemize}
    \item city: high-resolution (around $2560 \times 1920$, 1.85MB) image of a japanese market;
    \item animals: medium-resolution ($600 \times 225$, 338KB) poster with several animals;
    \item hubble: high-resolution image ($2300 \times 2100$, 6.90MB) from the Hubble telescope;
    \item bridge: Image of the San Francisco Golden Gate Bridge ($1165 \times 585$, 1.17MB);
    \item ImageNet: samples from ImageNet (using only CutMix);
    \item DeepInversion: synthetic images generated via DeepInversion~\citep{yinDreamingDistillDatafree2020};
    \item noise: static noise.
\end{itemize}

Images and samples are shown in Appendix \ref{apx:data_sources}. In general, we notice that using a single image is sufficient to reach a very high performance. However, there is still a large gap between the use of real data ($\data_i$) and a single out-of-distribution image. It is important to notice that DAC helps even when using the real data since we obtain an accuracy of $84.2$ against the $65.8$ of LwF. Moreover, even very different domains ("hubble") still work better than completely random data ("noise"). Finally, diversity, given either by a large $\data_{ood}$ or by heavy augmentations seems to be the most important factor since using ImageNet (more than 1M images) is slightly better than using data from the current task ($5000$ images).

\begin{table}
    \centering
    \caption{Test accuracy of DAC with different data sources on SplitCIFAR100 (10 Tasks).}\label{tbl:cifar100Abl}
    \scalebox{0.85}{
        \begin{tabular}{lcccr}
        \toprule
                      & \parbox{1cm}{Original \\ Domain}  & \parbox{1cm}{Single \\ Image} & Natural            & ACC \\ \midrule
        Current Data ($\data_i$)  & \checkmark &              & \checkmark         & \mustd{84.2}{0.5} \\
        ImageNet      &            &              & \checkmark         & \mustd{84.8}{0.6} \\
        animals       &            & \checkmark   & \checkmark         & \mustd{82.1}{0.5} \\
        city          &            & \checkmark   & \checkmark         & \mustd{80.5}{0.8} \\
        bridge        &            & \checkmark   & \checkmark         & \mustd{79.0}{0.6} \\
        hubble        &            & \checkmark   &                    & \mustd{65.1}{0.3} \\
        DeepInversion &            &              &                    & \mustd{64.9}{3.3} \\
        noise         &            & \checkmark   &                    & \mustd{10.7}{0.5} \\ \bottomrule
        \end{tabular}
        }
\end{table}

\subsection{Analysis}
We can summarize the main findings as follow:
\paragraph{DAC provides a good stability-plasticity tradeoff.} The separate consolidation step provides a better stability-plasticity tradeoff, higher accuracy, and better scaling w.r.t. the number of tasks, even in rehearsal-free scenarios. 
\paragraph{PLD Improves forward transfer between models.} In the sequential setting, DAC and the PLD loss shows positive forward transfer. This result suggests that latent distillation helps to combine models with different representations and to learn richer latent representations.  
\paragraph{Small data sources are sufficient.} Limited data sources, such as a single out-of-distribution image, show a good performance due to the heavy augmentations. This opens up to the opportunity of offloading the consolidation to a server without sharing the data.
\paragraph{Limitations} We had to limit the scope of our experimental analysis to two scenarios (the sequential and independent) even though most realistic scenarios will be a combination between the two. Furthermore, most CL methods are not applicable to DCL, which limits the number of baselines that we can fairly compare against. We compared against rehearsal-free baselines to mitigate this limitation.


\section{Related Works}

Recent work on knowledge distillation (KD)~\citep{hintonDistillingKnowledgeNeural2015} explores the possibility of distillation with limited data. Data-free KD is applied in the offline training setting by creating synthetic images with generators in ~\citep{liuDataFreeKnowledgeTransfer2021}. More relevant to our work, \citet{baradadjurjoLearningSeeLooking2021} propose handcrafted noise and simple procedurally generated images, while \citet{fangMosaickingDistillKnowledge2021} create image by combining together slices of several images. \citet{asanoExtrapolatingSingleImage2022} show the beneficial effect of heavy augmentations and \citet{beyerKnowledgeDistillationGood2022} additionally show the benefit of long training schedules.

In continual learning, many popular methods are based on knowledge distillation~\citep{liLearningForgetting2018,buzzegaDarkExperienceGeneral2020a}. Progress and Compress~\citep{schwarzProgressCompressScalable2018} uses an adaptation and compression step reminescent of our consolidation, but it still needs the original data. \citet{gomez-villaContinuallyLearningSelfSupervised2022} applies projected feature distillation in a continual self-supervised setting. Similar to us, they use a projection network to distill the features, however their method is purely unsupervised and neither can it handle distillation from multiple networks as we consider in this paper. 
In continual learning, exemplar-free scenarios are very popular, especially in the class-incremental setting (DFCIL). This scenario is different from our setting since the model still has access to $\data_i$ (see also Appendix \ref{apx:scenario_compare}). Many strategies address DFCIL by using alternative sources of data. \cite{cartaExModelContinualLearning2022} uses synthetic data generated via model inversion, \cite{zhangClassincrementalLearningDeep2020a} uses external data, and \cite{smithAlwaysBeDreaming2021} uses a data-free training process similar to GANs. \cite{leeOvercomingCatastrophicForgetting2019a} can optionally use external data for model calibration. \cite{9737321} and \cite{dongBridgingNonCooccurrence2021a} uses external data for semantic segmentation and object detection, respectively, by exploiting the notion of background on these tasks, which provide a neutral label, which makes them inapplicable for classification tasks. Among all these strategies, only \cite{cartaExModelContinualLearning2022} (ED, Table \ref{tab:res_core50}) and \cite{zhangClassincrementalLearningDeep2020a} (DMC, Table \ref{tbl:cifar100}) are fully applicable to DCL.

\section{Conclusion}
In this paper, we studied the problem of distributed continual learning (DCL), a scenario which provides two challenges: zero-cost transfer learning between SCDs and incremental learning from a stream of models. We proposed DAC as a general DCL method that satisfy the privacy and communication constraints. We showed that DAC reaches state-of-the-art performance using a single out-of-distribution image in rehearsal-free and distributed CL. Additionally, we DAC enables transfer beween SCDs.

DCL is a very general setting that opens up many research avenues. In this paper, we focused on the computational problem of consolidation, but many other aspects are worth of study, such as studying different communication protocols or sending other forms of encoded knowledge instead of the parameters. Knowledge consolidation also opens up the possibility of a continuous interaction between SCDs and large continually pretrained models.

To conclude, we would like to point out that improvements in DCL can easily transfer to the single-model/single-device scenario, as it happened for DAC, which improved the state-of-the-art performance on the task-incremental Split CIFAR100 and Split TinyImageNet. We hope that the research in distributed continual learning scenarios will provide insights about questions such as the relationship between adaptation and consolidation explored in this paper.

\bibliography{biblio}
\bibliographystyle{icml2023}

\newpage
\appendix
\onecolumn
\newpage

\section{Reproducibility}
We release our source code (anonymized in the supplementary material for the review, public on github afterwards). All of our experiments use Avalanche~\citep{lomonacoAvalancheEndtoEndLibrary2021}, a continual learning library based on PyTorch~\citep{paszkePyTorchImperativeStyle2019}. We release the experiments' configurations using Hydra~\citep{Yadan2019Hydra}(hierarchical yaml configuration files), which means that each experiment in the paper can be reproduced by running a main python script with the desired configuration, as detailed in the README of the source code. Experimental details relevant for independent implementations are available in the following sections.

\section{Comparison Between Related Learning Scenarios}\label{apx:scenario_compare}
The distributed continual learning scenario present some similarities with other scenarios in the literature. We provide a more detailed discussion of their differences here hoping to highlight their difference:
\begin{description}
    \item[continual learning]: a single model learning from a nonstationary stream of data. Knowledge sharing between models is not possible and current data is always available. Plasticity is suboptimal due to the stability/plasticity tradeoff.
    \item[rehearsal-free CL]: includes scenarios such as data-free class-incremental learning (DFCIL), where a single model learns from a nonstationary stream of data. Data from previous experiences is unavailable due to privacy constraints or severe storage limitations. Knowledge sharing is not possible and current data is always available. Plasticity is suboptimal due to the stability/plasticity tradeoff.
    \item[federated]: client-server organization with a single centralized controller. All the clients are learning the same task, possibly on different data. The server has full control over the training process and the client synchronize every few training iterations, resulting in multiple communication rounds. SCD data is private but plasticity is suboptimal because SCD are optimized on the global instead of the local task.
    \item[DCL]: Each SCD learn its own task, keeping all the data private. Plasticity is optimal because SCD are optimizing their own task. Communication is minimized by using a single round of communication.
\end{description}

Figure \ref{fig:scenario_full} shows the training process of the four different scenarios, mapping the adaptation and consolidation phases defined in Section \ref{sec:scenario} to the other scenarios. Table \ref{tbl:scenario_properties} summarizes the properties of the different scenario.

\begin{table}[h]
    \centering
    \caption{Summary of the main properties of different scenarios related to distributed continual learning.}\label{tbl:scenario_properties}
    \begin{tabular}{lrrrrr}
        \toprule
         & \thead{Past Data \\ Privacy} & \thead{Current Data \\ Privacy} & \thead{Optimal \\ plasticity} & \thead{Multiple \\ Devices} & \thead{single round \\ of communication} \\
        \midrule
        Continual Learning &  & & & & \\
        Rehearsal-free Continual Learning & \checkmark & & & \\
        Federated Learning & \checkmark & & & \checkmark & \\
        Ours & \checkmark & \checkmark & \checkmark & \checkmark & \checkmark\\
        \bottomrule
    \end{tabular}
\end{table}

\begin{center}
    \includegraphics[width=0.9\textwidth]{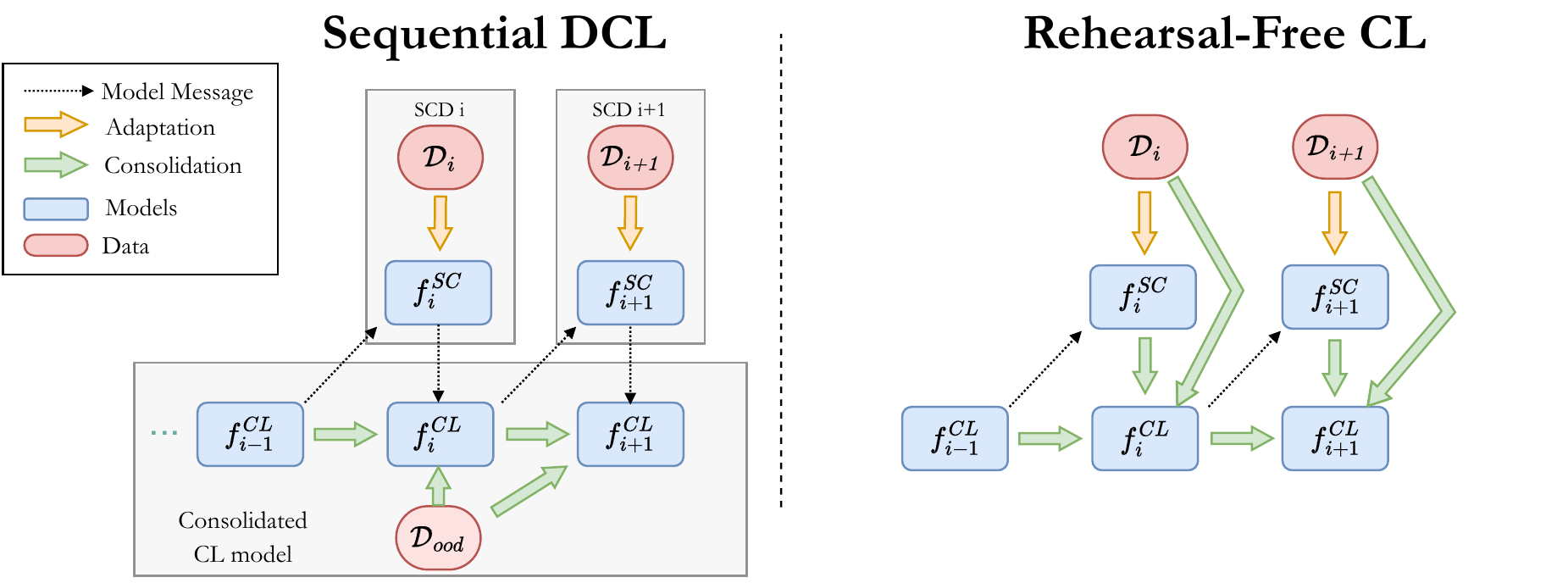}
    \captionof{figure}{Schematic comparison of Sequential DCL vs Rehearsal-free CL, showing a possible implementation of DAC in rehearsal-free. Notice that in rehearsal-free CL we have the possibility to use the original data during consolidation and all the learning steps can happen on the same device.}\label{fig:scenario_full}
\end{center}

\section{Data Sources}\label{apx:data_sources}
In this section, we show samples from the images used for knowledge distillation.
\emph{city}: high-resolution (around 2560x1920, 1.85MB) image of a japanese market;

\begin{minipage}{0.45\textwidth}
    animals: medium-resolution (600x225, 338KB) poster with several animals;
    
    \includegraphics[width=0.4\textwidth]{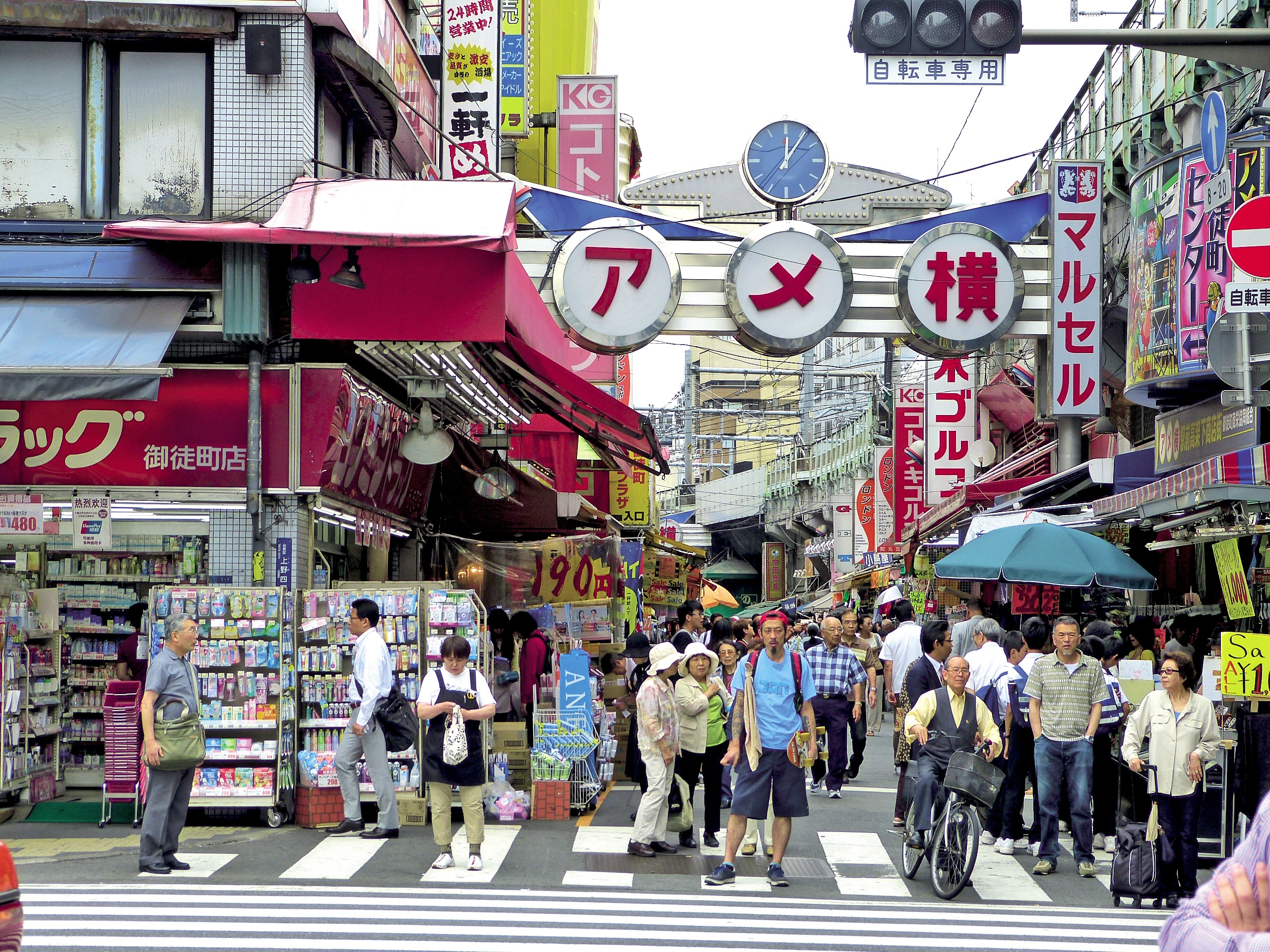}
    \includegraphics[width=0.1\textwidth]{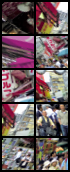}
    
    hubble: high-resolution image (2300x2100, 6.90MB) from the Hubble telescope;
    
    \includegraphics[width=0.4\textwidth]{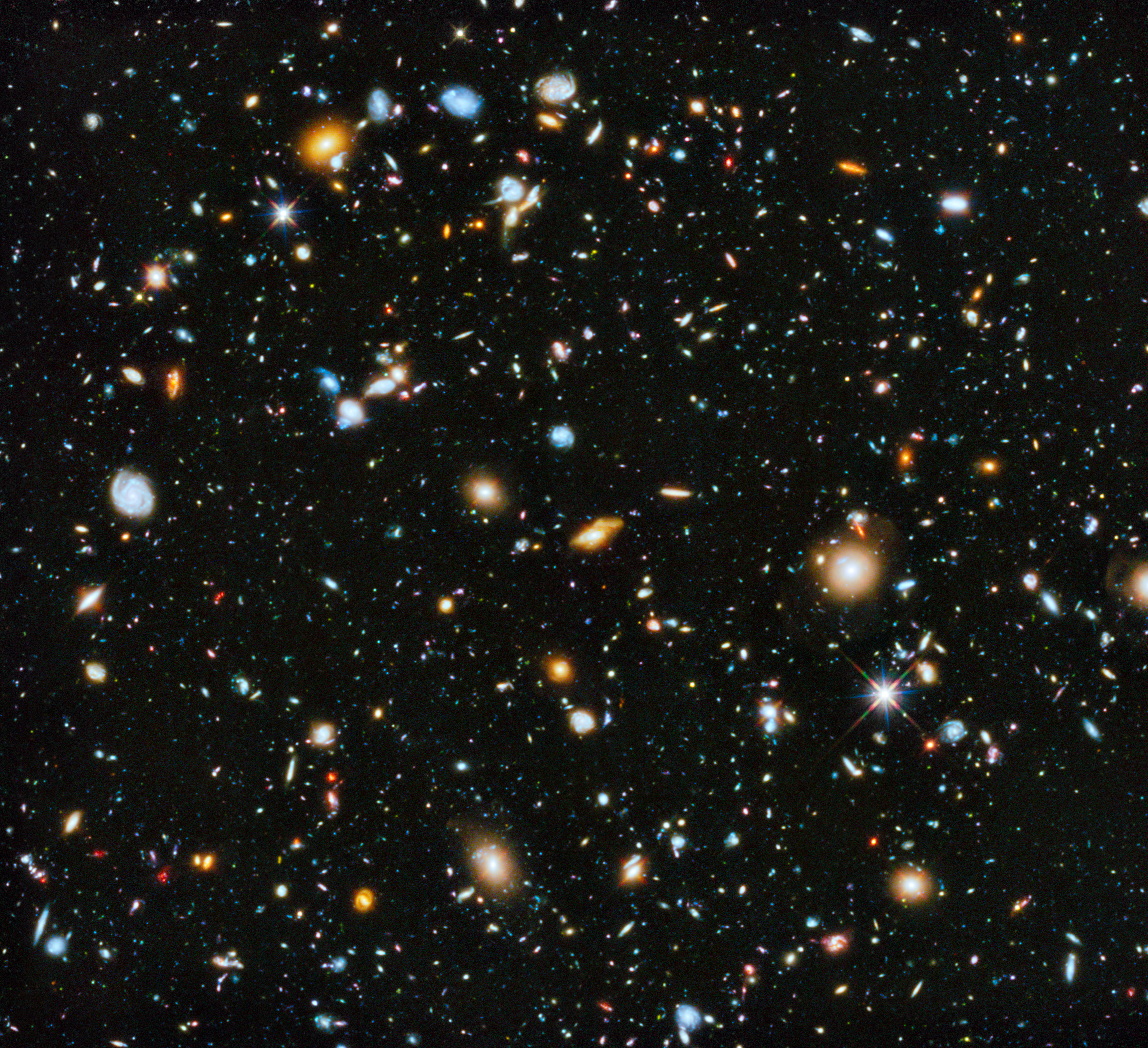}
    \includegraphics[width=0.1\textwidth]{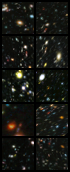}
    
    animals: medium-resolution (600x225, 338KB) poster with several animals;
    \includegraphics[width=0.4\textwidth]{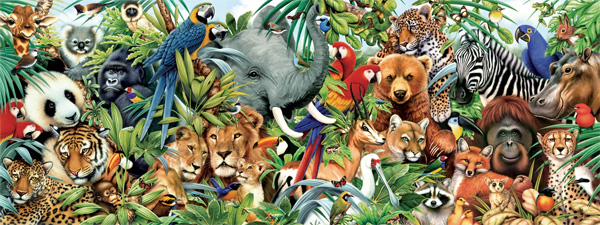}
    \includegraphics[width=0.1\textwidth]{imgs/aug/ameyoko.png}

\end{minipage}
\begin{minipage}{0.45\textwidth}
    bridge: Image of the San Francisco Golden Gate Bridge (1165x585, 1.17MB);
    
    \includegraphics[width=0.4\textwidth]{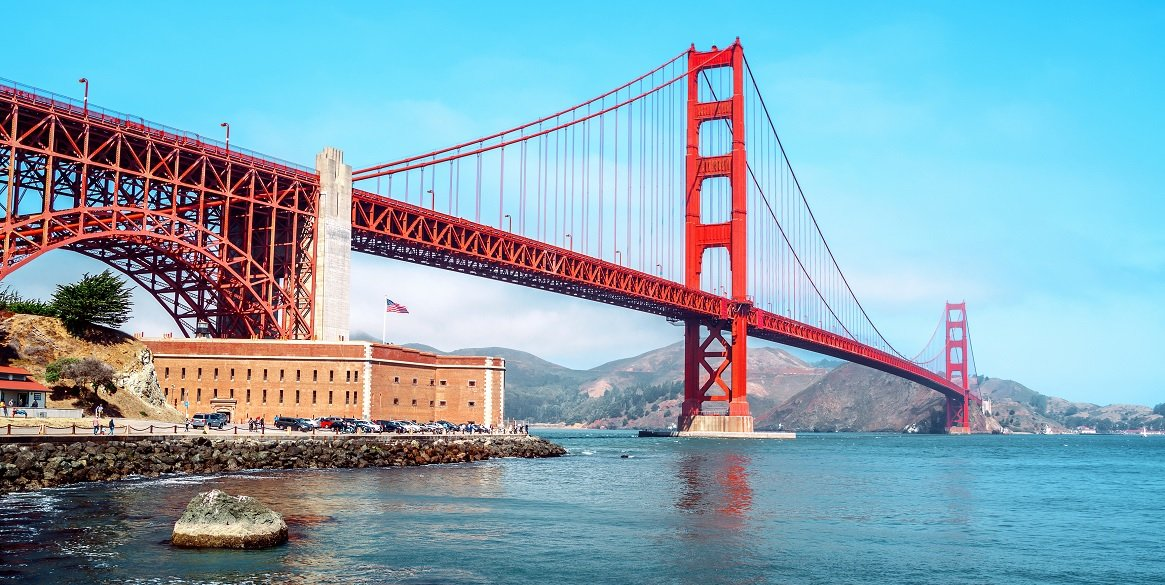}
    \includegraphics[width=0.1\textwidth]{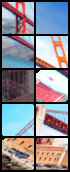}
    
    ImageNet: samples from ImageNet (without augmentations);
    
    \includegraphics[width=0.1\textwidth]{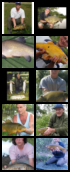}
    
    noise: static noise.
    
    \includegraphics[width=0.3\textwidth]{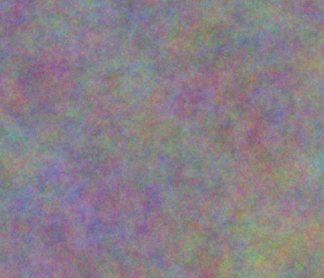}
    \includegraphics[width=0.1\textwidth]{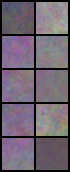}
\end{minipage}

\section{Hyperparameters}\label{apx:hparams}
\paragraph{SplitCIFAR100:} We use a slimmed ResNet18 as a backbone for both the teacher and consolidated model.
During the consolidation, we use Adam with learning rate set to $0.0001$, with a batch size of $512$ and $500'000$ iterations. We use a temperature of $0.5$ for distillation. For the PLD loss, we set $\lambda=0.01$ and apply the loss at \verb|layer4.0| and \verb|linear| (logits). 

\paragraph{CORe50:} We use a MobileNet v2 pretrained on ImageNet as a backbone for both the teacher and consolidated model.
During the consolidation, we use Adam with learning rate set to $0.0001$, with a batch size of $128$ and $100'000$ iterations. We use a temperature of $1.0$ for distillation. For the PLD loss, we set $\lambda=100.0$ and apply the loss at \verb|classifier| (logits). 

\section{CKA}\label{apx:cka}

In this section, we show the CKA, as described in Section \ref{sec:cka_forward}, for the entire stream. We compute the CKA between the first expert and the expert after experience $i$.



\begin{center}
    \includegraphics[width=\textwidth]{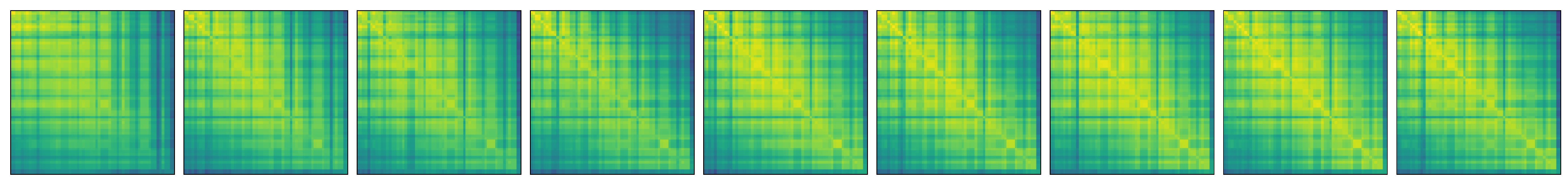}
    \captionof{figure}{CKA for \nopld.}
\end{center}

\begin{center}
    \includegraphics[width=\textwidth]{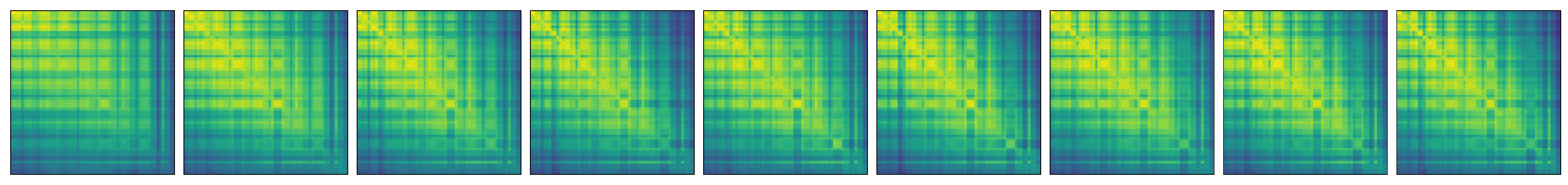}
    \captionof{figure}{CKA for \expdac.}
\end{center}


\end{document}